\documentclass[]{spie}  %>>> use for US letter paper
\usepackage{amsmath,amsfonts,amssymb}
\usepackage{graphicx}
\usepackage[colorlinks=true, allcolors=blue]{hyperref}
\usepackage[capitalise]{cleveref}
\usepackage{cite}
\usepackage{algorithmic}
\usepackage{textcomp}
\usepackage{xcolor}
\usepackage{algorithmic}
\usepackage[labelsep=period]{caption}
\usepackage{cuted}
\usepackage{array}
\usepackage{capt-of}
\usepackage[caption=false,font=footnotesize]{subfig}
\usepackage{comment}
\usepackage{caption}
\usepackage{lineno}
\usepackage{cleveref}
\usepackage{multirow}
\usepackage{floatrow}

\crefname{equation}{Eq.}{Eq.} % capitalize "E", no period
\crefrangelabelformat{equation}{(#3#1#4)--(#5#2#6)}
\crefname{figure}{Fig.}{Fig.}

\def\BibTeX{{\rm B\kern-.05em{\sc i\kern-.025em b}\kern-.08em
    T\kern-.1667em\lower.7ex\hbox{E}\kern-.125emX}}

\title{	Memory capacity analysis of time-delay reservoir computing based on silicon microring resonator nonlinearities}

\author[a]{Bernard J. Giron Castro}
\author[b]{Christophe Peucheret}
\author[a]{Francesco Da Ros}
\affil[a]{DTU Electro, Technical University of Denmark, Ørsteds Plads, 2800 Kgs. Lyngby, Denmark}
\affil[b]{Univ Rennes, CNRS, UMR6082 - FOTON, 22305 Lannion, France}

\authorinfo{Bernard J. Giron Castro: E-mail: bjgca@dtu.dk}

\begin{document} 
\maketitle

\begin{abstract}
Silicon microring resonators (MRRs) have shown strong potential in acting as the nonlinear nodes of photonic reservoir computing (RC) schemes. By using nonlinearities within a silicon MRR, such as the ones caused by free-carrier dispersion (FCD) and thermo-optic (TO) effects, it is possible to map the input data of the RC to a higher dimensional space. Furthermore, by adding an external waveguide between the through and add ports of the MRR, it is possible to implement a time-delay RC (TDRC) with enhanced memory. The input from the through port is fed back into the add port of the ring with the delay applied by the external waveguide effectively adding memory. In a TDRC, the nodes (virtual) are multiplexed in time, and their respective time evolutions are detected at the drop port. The performance of MRR-based TDRC is highly dependent on the amount of nonlinearity in the MRR. The nonlinear effects, in turn, are dependent on the physical properties of the MRR as they determine the lifetime of the effects. Another factor to take into account is the stability of the MRR response, as strong time-domain discontinuities at the drop port are known to emerge from FCD nonlinearities due to self-pulsing (high nonlinear behaviour). However, quantifying the right amount of nonlinearity that RC needs for a certain task in order to achieve optimum performance is challenging. Therefore, further analysis is required to fully understand the nonlinear dynamics of this TDRC setup. Here, we quantify the nonlinear and linear memory capacity of the previously described microring-based TDRC scheme, as a function of the time constants of the generated carriers and the thermal of the TO effects. We analyze the properties of the TDRC dynamics that generate the parameter space, in terms of input signal power and frequency detuning range, over which conventional RC tasks can be satisfactorily performed by the TDRC scheme.
\end{abstract}


\section{Introduction}
\label{sec:intro}  % \label{} allows reference to this section

 The current rise of artificial intelligence and neural network algorithms has demonstrated their potential applications in every scientific discipline and everyday life. Current neural network algorithms, while powerful, require a large number of computing resources for their often complex training procedures\cite{Schuman2022, Huang2022, Cucchi_2022}. Therefore, more powerful and sophisticated computing architectures with lower power consumption are currently under research. Neuromorphic computing promises to push forward the current boundaries of conventional computing. Some of its potential features are massive parallel computing, collocation of the memory, and the computing processor which avoids the so-called Von Neumann bottleneck that limits the achievable throughput of current computing hardware \cite{Schuman2022, Huang2022, Cucchi_2022}. 
 
Under the described context, we focus on reservoir computing (RC). RC is a type of neuromorphic computing approach, based on recurrent neural networks, which is characterized by the capability of solving complex computational tasks while requiring a relatively simple and fast training process of the output layer \cite{Schuman2022, Huang2022, Cucchi_2022}. The data at the input layer is mapped to a higher dimensional reservoir state through nodes interconnected with random and fixed weights. The nodes in the reservoir layer provide the nonlinear dynamics and complexity required to accurately address computational tasks. However, if the influence of the dynamics is too strong, it might deter any influence of the input in the reservoir state. The effect of the current and past input on the reservoir response at a given instant is essential in computing tasks that require memory. Nevertheless, the buffering of past inputs must be finite, as otherwise, it would considerably increase the computing load of RC. Hence, such conditions set two of the main features of a functional RC: First, a certain degree of nonlinearity so that the input data is projected into a high-dimensional reservoir state following a nonlinear temporal expansion. This expansion allows RC to linearly differentiate the RC states to address them through linear regression at the output layer. Secondly, the fading memory\cite{jaeger2001short}, by which the input signals of the reservoir keep propagating over a finite time through the nodes of the reservoir. Consequently, a temporary dependence of the current RC state on previous inputs is formed. Other factors, such as the number of nodes, the impact of neuron biases, and the sparsity are also important in RC, although they are not the main focus of this study \cite{Cucchi_2022}. 

Among the approaches to physical RC, photonics has demonstrated some interesting implementations through a variety of different photonics devices, such as Mach-Zehnder modulators in a optoelectronic loop\cite{Paquot2012}, laser sources \cite{Bueno:17, 8758193, Skontranis_2023}, semiconductor optical amplifiers\cite{Duport:12} and microring resonators \cite{doi:10.34133/icomputing.0067}, etc. The dynamics of those devices have demonstrated the potential to provide the aforementioned benefits of neuromorphic computing.  Some of these works minimize the number of nonlinear nodes required by the scheme using time multiplexing of a single physical nonlinear node in a method called time-delay RC (TDRC). Each virtual node is processed through the nonlinear node, one at a time, and the RC output is determined after all the virtual nodes are sampled. Several photonic TDRC implementations are found in the literature\cite{Duport:12, Paquot2012, Chen:19}. MRR-based TDRC schemes have achieved high performance in diverse time-series prediction tasks in previous studies \cite{doi:10.34133/icomputing.0067, GironCastro:24}. The conventional approach of the scheme, further detailed in \cref{sec:2}, uses an external waveguide connected between the through and add ports of an add-drop MRR to provide delayed feedback of the input and enhance the memory of the system \cite{doi:10.34133/icomputing.0067}. 

The required nonlinear dynamics are triggered in the MRR by two-photon absorption (TPA). Due to TPA, an excess of free carriers emerges in the MRR, which in turn produces free-carrier dispersion (FCD). The free carrier absorption within the cavity generates heat. Both the resulting FCD and the thermo-optic (TO) effect change the refractive index of the silicon MRR waveguide. This causes a nonlinear shift of the cold cavity resonance where FCD causes a blue shift and the TO effect a red shift, leading to opposite shifts of the resonance and an oscillatory self-pulsing of the cavity optical modal amplitude. \cite{doi:10.34133/icomputing.0067, PhysRevA.87.053805}. However, FCD and the TO effect usually differ in the order of magnitude of their timescales. The timescale of FCD is determined by the free carrier lifetime, $\tau_{\textrm {FC}}$ while the timescale of the thermal effects is related to the thermal diffusion time constant $\tau_{\textrm {th}}$. It is better to work with optical pulses modulated by a sequence with a symbol rate closer to the lowest of the two timescales, i.e., $\tau_{\textrm {FC}}$ to increase the TDRC processing speed \cite{doi:10.34133/icomputing.0067}.  Recently, we studied the effects of varying $\tau_{\textrm {FC}}$, $\tau_{\textrm {th}}$, as well as the waveguide losses, on the performance of MRR-based TDRC \cite{GironCastro:24}. The results showed a high correlation, in terms of the timescales of the nonlinear processes, between the prediction performance and the transitions from linear to nonlinear regimes of the MRR. The regime of low levels of nonlinearity and the regime of self-pulsing in the MRR were both found to be detrimental to the performance of the RC while the best performance was found in a narrow regime between them. Nevertheless, such findings were limited to the NARMA-10 task.

Here, we numerically deepen the analysis of the memory and nonlinearity properties of silicon MRR-based TDRC and relate it to the insights obtained for the NARMA-10 task \cite{GironCastro:24}. We also focus on the role that both nonlinearity and memory play in this scheme when addressing a variety of RC conventional computing tasks with different applications. In this way, a better understanding of the properties of MRR-based TDRC can be achieved, which allows further generalization of the insights obtained in the previous study\cite{GironCastro:24}, to other tasks. The manuscript is structured as follows: \cref{sec:2} describes the working mechanism of the TDRC scheme analyzed in this work as well as relevant insights from previous studies. \cref{sec:3} outlines the memory capacity metrics used in this article. Additionally, we analyze the existence in this scheme of the (potentially) universal trade-off between memory and nonlinearity as previously investigated in the literature and for the studied TDRC. The results of the study are presented in \cref{sec:4} and discussed in further detail in \cref{sec:5}. \cref{conclusion} summarizes the main conclusions of this work.

\section{Silicon MRR-based TDRC}\label{sec:2}

\begin{figure}[t!]
\centering
    \includegraphics[scale=0.33, trim={0cm 0.2cm 0cm 0cm}]{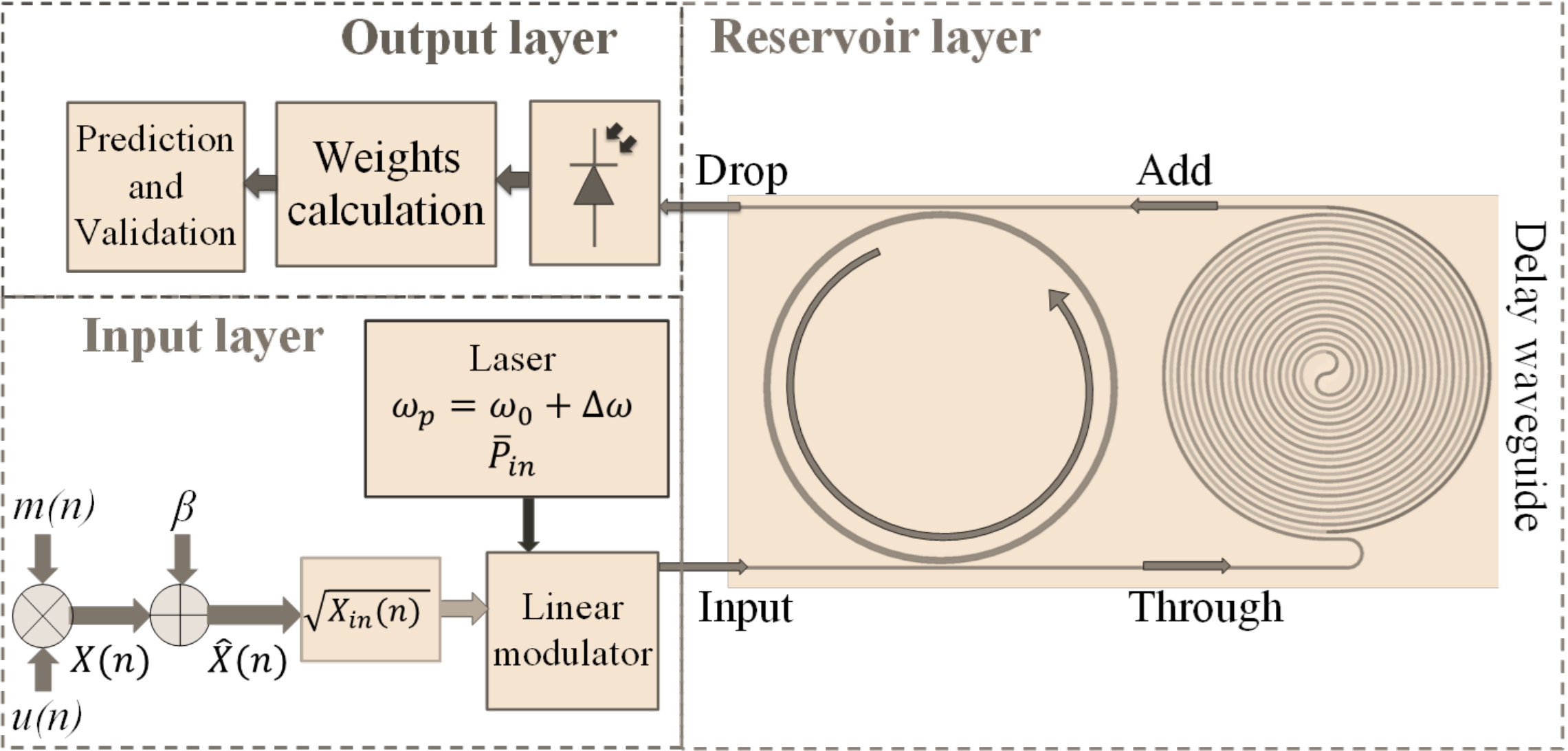}
\captionof{figure}{MRR-based TDRC setup scheme under investigation.}
\vspace{-0.1cm}
\label{fig1}
\end{figure}

The system under investigation is depicted in \cref{fig1}. The input data sequence $u(n)$, at a 1.0 GBd symbol rate, is multiplied by a masking sequence $m(n)$, which is generated from a uniform distribution in the range [0.0, 1.0]. In the masked input signal, $X(n)$, we set the random and fixed weights of each virtual node. Therefore, the mask sequence length must be equal to the desired number of nodes, $N$. Throughout this work, $N = 50$. An optimized bias $\beta$ is added to the masked input, forming $\hat{X}(n)$. The electric signal is linearly modulated on a continuous wave signal generated by a laser source. The optical signal has an average power $\overline{P}_{\textrm {in}}$ and an angular frequency $\omega_\textrm p$. The value of $\omega_\textrm p$ is detuned by $\Delta\omega$ from the resonance frequency of the MRR, $\omega_\textrm 0$. A root square function accounts for the optoelectronic transformation performed in the simulation as the model is based on the electric field at the input port. Hence, $X_{\textrm{in}}(n)$ denotes the resulting linearly modulated optical power at the input port. As aforementioned, the through and add ports of the MRR are linked by an external waveguide which adds a delay of $\tau_\textrm d = 0.5$ ns to the optical signal that propagates to the add port. The linear combination of the electric fields coupling from the cavity and the add port is detected by a photodiode. The output electrical signal is post-processed to sample the virtual nodes and train the weights of the output layer by means of ridge regression. The prediction of the TDRC is averaged over 10 different seeds used to generate the original input sequences of each addressed computing task. Comprehensive details regarding the simulation model of the system as well as the optical parameters used for the simulations are presented in our aforementioned study \cite{GironCastro:24}.

\section{Information Processing Capacity in RC}\label{sec:3}

It is important to introduce some useful memory-related metrics to quantify the memory capacity of the studied system. On the other hand, memory is not always a very relevant feature for some computational tasks and may even be detrimental if it counters the nonlinear features of the reservoir. Furthermore, the performance of memory-demanding tasks can be compromised if the nonlinearity of the reservoir is too high. Because of this trade-off between nonlinearity and memory \cite{Inubushi2017} it is important to also introduce a relevant metric for the nonlinearity of this type of photonic TDRC. This allows us to establish a clearer picture of the balance between the memory and nonlinear features.

\subsection{Linear, nonlinear and total memory capacity}\label{subsec:3.1}

The fading memory of a reservoir entails that it cannot solve unbounded-time memory tasks and therefore, the number of previous inputs in the past that it can accurately learn is limited. Thus, it is possible to quantify the memory capacity of RC within a finite upper bounded range. The linear memory capacity, as traditionally defined\cite{jaeger2001short}, is a measurement of the ability of RC schemes to reconstruct a target function defined by previous inputs of the RC. In other words, the task of the RC is to reconstruct the input (usually uniformly distributed) $k$ steps in the past, $y_k(n) = u(n-k)$. The capacity for a specific value of $k$ is then\cite{jaeger2001short}:

    \begin{equation}\label{eq1}
        C[y_k] = 1-\textrm{NMSE}[y_k], 
    \end{equation}

    \begin{equation}\label{eq2}
    \textrm {NMSE} = \frac{1}{L_{\textrm {data}}}\frac{\sum_{n} \left( \hat{y}(n) - y(n) \right)^2}{\sigma_{y}^2},    
    \end{equation}

where NMSE is the normalized mean square error, $\hat{y}(n)$ and $y(n)$ are the estimated and target sequences, respectively, with a length $L_{\textrm {data}}$. $\sigma_{y}^2$ is the standard deviation of $y(n)$. It has been demonstrated\cite{Dambre2012} that the maximum number ($k_{\textrm{max}}$) of significant $k$ steps contributing to the memory capacity, is upper-bounded by the number of nodes. Hence, for this work $k_{\textrm{max}} = 50$. 
Subsequently, the total linear memory capacity is defined as the sum of the capacities over all values of $k\leq k_{\textrm{max}}$:

\begin{equation}\label{eq3}
    C_{\textrm{lin}} = \sum_k^{k_{\textrm{max}}}{C[y_k]}. 
\end{equation}

A generalization of the memory capacity for higher-order functions is possible by considering the reconstruction of a set of basis functions in the Hilbert space of fading memory functions\cite{Dambre2012, HülserKöster2023}. Sometimes referred to as nonlinear memory capacity, it is determined by considering a basis out of finite products of higher-order Legendre polynomials. For example, the second-order memory capacity, referred to as quadratic memory capacity\cite{Duport:12}, is reconstructed using the second-order Legendre polynomial as the target function: $y_k(n) = 3u^2(n-k)-1$. The basis function of the $i^{\textrm{th}}$-order is denominated $\mathbb{P}_i$, so that $y_i(n) = \mathbb{P}_i$. Thus, the nonlinear capacities can be expressed in a general expression as:

\begin{equation}\label{eq4}
    C_i = \sum_k^{k_{\textrm{max}}}{1-\textrm {NMSE}[\mathbb{P}_i,  \hat{y}_i(n)]}. 
\end{equation}

Then, the total memory capacity $MC$, also referred to as information processing capacity, \cite{jaeger2001short, HülserKöster2023} is the sum of all capacities of the $i^{\textrm{th}}$ order that are calculated up to the highest order of function considered, $M$:

    \begin{equation}\label{eq5}
        MC = \sum_i^{M}{C_i}. 
    \end{equation}

\subsection{Nonlinearity metric of the MRR-based TDRC}\label{subsec:3.2}

The coefficient of determination $R^2$ was used in the previous study of the system\cite{GironCastro:24} in an attempt to quantify the nonlinear transformation of the input data at the drop port of the MRR before being affected by the nonlinearity given by the photodiode. This metric provided some insight into how the output states of the RC could be fully accounted as a linear transformation of the input data (a value close to 1.0), if there is no direct relation between them (value of 0.0), or somewhere in between (where the best performance was obtained).

Here, we use a metric more closely related to the physics involved in the dynamics of the MRR by quantifying the frequency detuning of the resonance resulting from the nonlinear effects occurring in the MRR, named $\delta_{\textrm{NL}}(t)$. This term is hence, the sum of the frequency detuning due to FCD and the one due to the TO effect\cite{GironCastro:24}: 

\begin{equation}\label{eq6}
\delta_{\textrm{NL}}(t) = \frac{1}{2\pi}\left[\Delta\omega_{\Delta N}(t) + \Delta\omega_{\Delta T}(t)\right].
\end{equation}

Nevertheless, the value of $\delta_{\textrm{NL}}(t)$ is dependent on the instant value of the data sequence. Consequently, we calculate the standard deviation of $\delta_{\textrm{NL}}(t)$, i.e. $\sigma(\delta_{\textrm{NL}}(t))$, over the total length of the sequence. This gives a clearer insight into the strength of the nonlinear effects considered in the MRR for the processed data sequence.

\subsection{Relation between memory capacity, nonlinearity, and task performance}\label{subsec:3.3}

The values of each order of capacity and $MC$ give us a general idea of the memory capabilities of a particular RC scheme. Nonetheless, the translation of memory capacity to task performance is not straightforward. In principle, memory might not be the main feature required by RC to accurately address a specific task. Other tasks may be more reliant on the nonlinearity and dimensionality of the scheme (kernel quality\cite{Cucchi_2022}). Previous works\cite{Inubushi2017, Dambre2012} have demonstrated an apparent universal trade-off between nonlinearity and memory capacity in dynamical systems that require nonlinear transformations of the input to achieve a dimensionality expansion. Therefore, the memory capability of RC is ultimately limited by the amount of nonlinearity required in a specific scheme and task. Additionally, it has been shown that in some RC implementations, the total memory capacity might be poorly correlated with the performance of a specific task \cite{HülserKöster2023}, which is an important insight to take into account in this analysis.  Furthermore, different orders of memory capacity can have different computing significance when addressing a particular task, depending on the mathematical definition of each task. All around, it is evident the many factors to consider when analyzing the memory capabilities of the system under investigation.

\section{Results}\label{sec:4}
\subsection{MRR-based TDRC tasks and benchmarks}\label{subsec:4.1}

First, we introduce in this subsection a set of computing tasks, that have been previously defined and addressed with high performance in photonic TDRC implementations, and that are considered in this work \cite{Paquot2012, Duport:12, Chen:19, doi:10.34133/icomputing.0067}. Each of them has different performance metrics and diverse requirements of memory and nonlinearity. Thus, we verify if there is a common behaviour between their performances based on the nonlinear and memory dynamics of the MRR-based TDRC. It is important to note that each of the performance results could be further improved by an optimization of the optical parameters of the modulated optical signal, which is outside the main scope of this work. The reference task for this work, based on the previous analysis of this system \cite{GironCastro:24}, is the discrete-time tenth-order nonlinear auto-regressive moving average (NARMA-10). We present the results in terms of performance metric, of varying $\tau_{\textrm {FC}}$ in each of the implemented computing tasks for a range of $\overline{P}_{\textrm {in}} = [-20: 20]$ dBm and a $\Delta\omega/2\pi$ range of = $\pm 300$ GHz.

\subsubsection{NARMA-10 task}

NARMA-10 is a one-step ahead time-series prediction where the computing target is the function expressed as:

\begin{equation}
\label{eq7}
    y(n+1) = 0.3y(n) + 0.05y(n)\left[\sum_{i=0}^9 y(n-i)\right] + 1.5u(n-9)u(n) + 0.1 . 
\end{equation}

The sequence $u(n)$ is generated from a uniform distribution over the interval [0.0, 0.5]. The mask $m(n)$ is taken from a uniform distribution in the range [0.0, +1.0]. The performance is measured by calculating the NMSE between the predicted and target sequences as defined in \cref{eq2}. 2000 data points were used for training and 2000 for testing. Each of the analyzed tasks uses a warm-up sequence of 200 data points before both the training and testing sets to mitigate the effects of the initial state of the TDRC. 

\begin{figure}[h!]
\centering
    \includegraphics[scale=0.43, trim={0.0cm 4.2cm 0cm 4.2cm}]{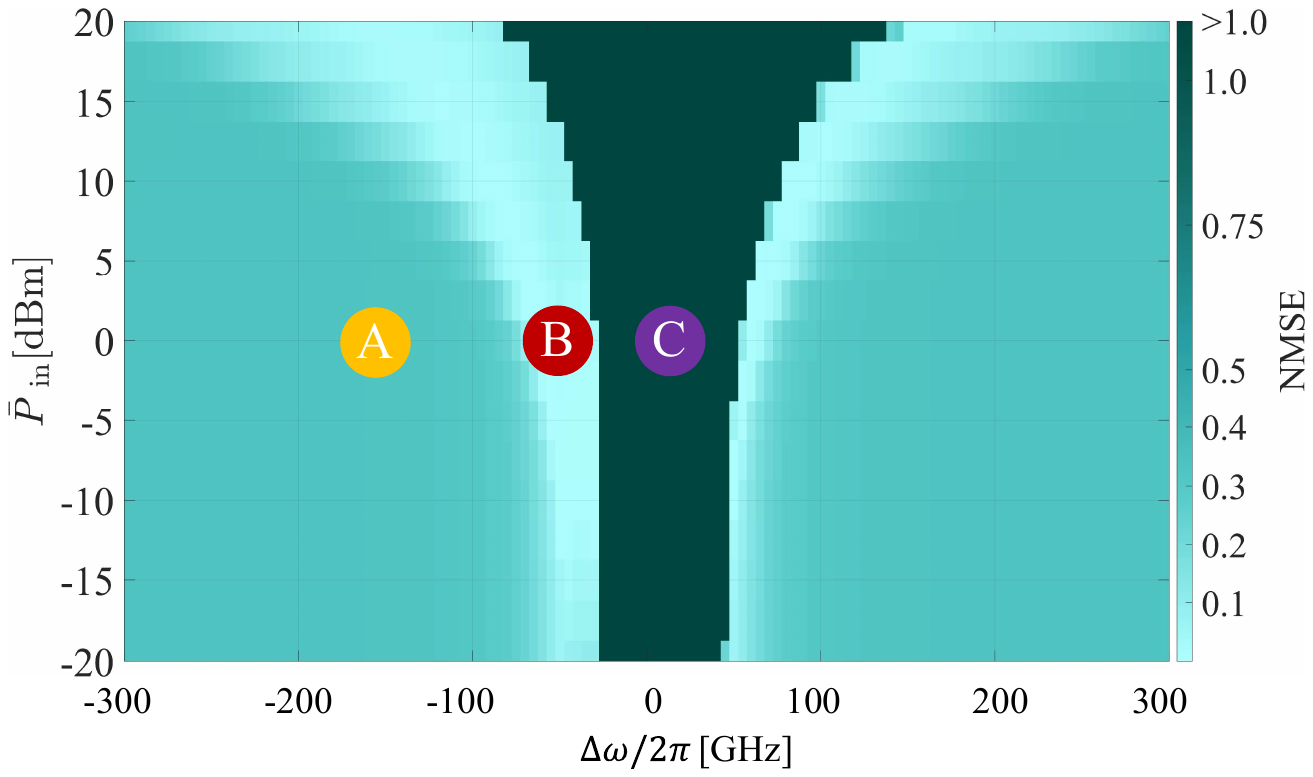}
\captionof{figure}{Regions A, B, and C in terms of $\overline{P}_{\textrm {in}}$ and $\Delta\omega/2\pi$, when solving the NARMA-10 task.}
\label{fig2}
\end{figure}

\begin{figure}[h!]
\centering
    \includegraphics[scale=0.617, trim={0.0cm 12.0cm 0cm 4.2cm}]{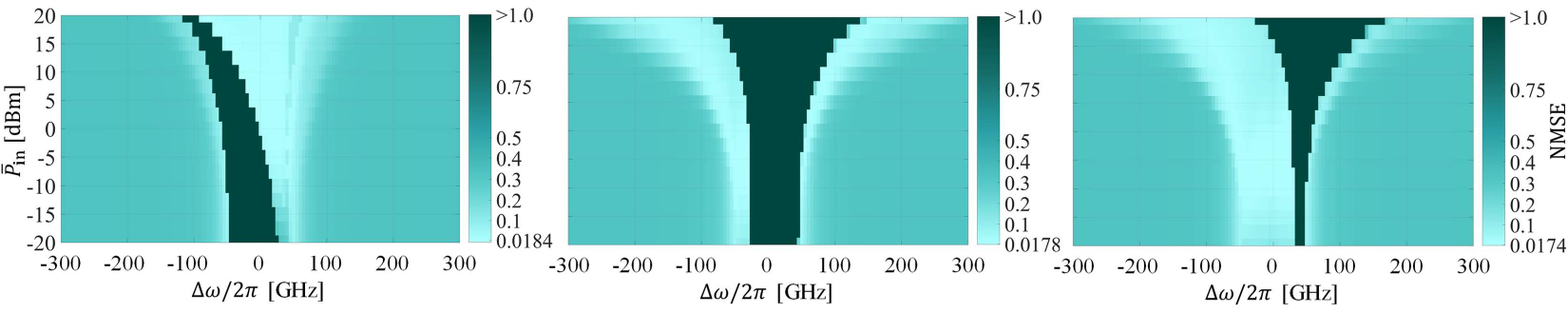}
\captionof{figure}{TDRC NMSE when solving the NARMA-10 task as a function of $\overline{P}_{\textrm {in}}$ and $\Delta\omega/2\pi$  for a $\tau_{\textrm {FC}}$ value of: a) 10 ps, b) 10 ns, c) 25 ns. Minima and maxima values are displayed as extreme values of the colorbar legend.}
\label{fig3}
\end{figure}

Three regions of different levels of performance were identified\cite{GironCastro:24} in the parameter space of $\overline{P}_{\textrm {in}}$ vs $\Delta\omega$ when the MRR-based TDRC addresses the NARMA-10 task. The extension of each regime highly depends on $\tau_{\textrm {FC}}$ and $\tau_{\textrm {th}}$. An instance of the results showing those regions (labeled A, B, C) is shown in \cref{fig2} for $\tau_{\textrm {FC}}$ = 10 ns and ${\tau_{\textrm {th}} = 50\ \textrm{ns}}$ with a waveguide attenuation of 0.8 dB/cm. Region A corresponds to an approximately linear regime of the MRR cavity in which the NMSE value of the prediction is upper-bounded by the contribution of nonlinearity given by the photodiode. In region B, an adequate level of nonlinearity is present in the system and the best performance is achieved (NMSE $<$ 0.2). Region C is characterized by the presence of self-pulsing, a nonlinear phenomenon in which the fast oscillations of the optical signal highly disturb the prediction performance of the system, making it completely unable to address the task (NMSE $>$ 1.0). By varying $\tau_{\textrm {FC}}$ to 10 ps and 25 ns, we can observe how the defined regions are changed. The parameter space of region B is extended while improving the prediction performance of the system\cite{GironCastro:24} (\cref{fig3}). 

\subsubsection{IPIX radar task}

\begin{figure}[h!]
\centering
    \includegraphics[scale=0.615, trim={0.0cm 6.5cm 0cm 4.0cm}]{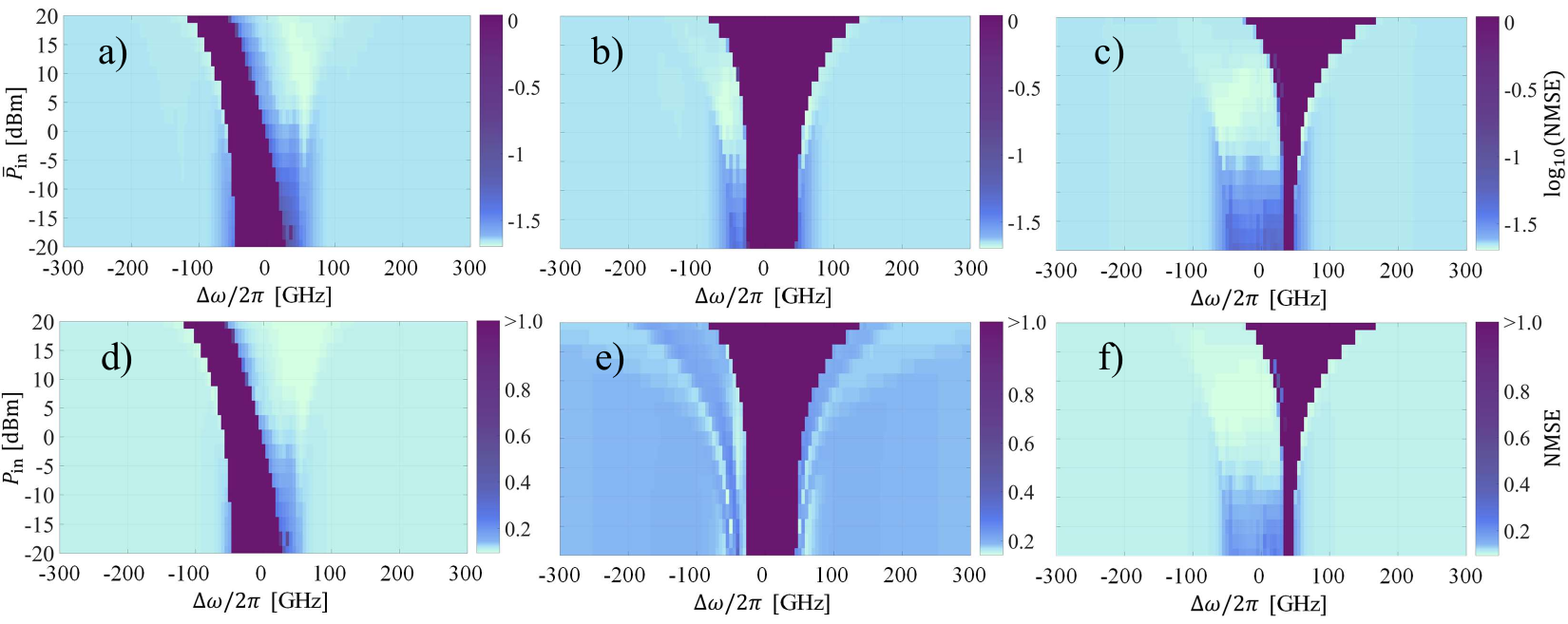}
\captionof{figure}{NMSE of the TDRC when solving the radar task as a function of $\overline{P}_{\textrm {in}}$ and $\Delta\omega/2\pi$  for a $\tau_{\textrm {FC}}$ value of: 10 ps (a, d), 10 ns (b, e) and 25 ns (c, f), with $k = 1$ (a - c) and $k = 2$ (d - f).}
\label{fig4}
\end{figure}

An experimental backscattered radar signal from the ocean surface is used in this prediction task. The signal was measured by the McMaster University IPIX radar \cite{Haykin}. There are two data sets available, which refer to the low (average wave height of 0.8 meters) and high (average wave height of 1.8 meters) sea states. We focus on the high sea states dataset in this study. The target of the task is the measured signal, shifted $k$ steps in the future. Therefore, we calculate the NMSE between the prediction and the actual data of the signal. The dataset contains the in-phase and in-quadrature outputs of the radar demodulator. Hence, the 2-D signal is first flattened and then processed sequentially by the system. We use 1000 samples for the training and 1000 for the testing. We simulated the system for $k = 1$ and $k = 2$. The performance of the TDRC when addressing the radar task as a function of $\overline{P}_{\textrm {in}}$ and $\Delta\omega$ is shown in \cref{fig4} for different values of $\tau_{\textrm{FC}}$. \cref{fig4} a) - c) correspond to $k = 1$ and d) - f) to $k = 2$. For visualization purposes, we present the results for $k = 1$ as the base-10 logarithm of the NMSE.   

\subsubsection{Signal classification}

\begin{figure}[h!]
\centering
    \includegraphics[scale=0.62, trim={0.0cm 12.0cm 0cm 4.2cm}]{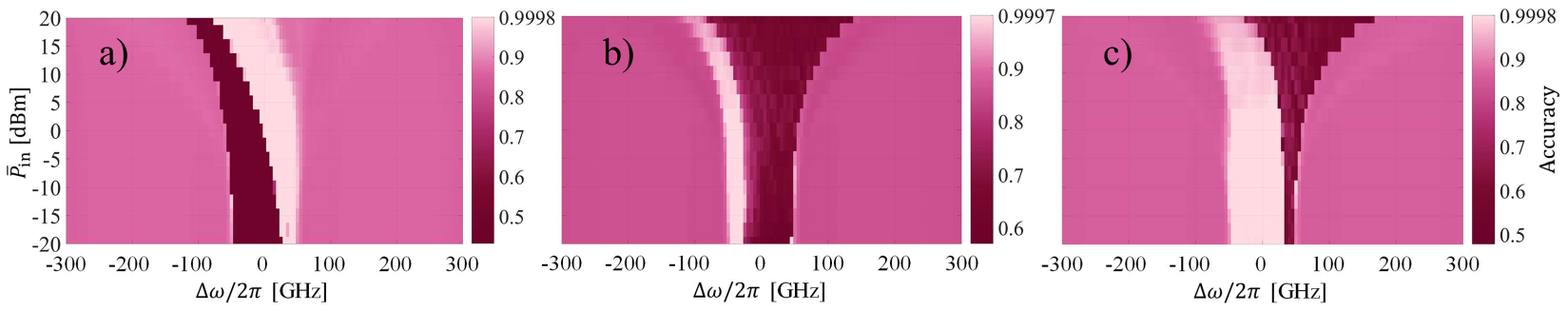}
\captionof{figure}{Accuracy of the TDRC when solving the signal classification task as a function of $\overline{P}_{\textrm {in}}$ and $\Delta\omega/2\pi$ for a $\tau_{\textrm {FC}}$ value of: a) 10 ps, b) 10 ns, c) 25 ns.}
\label{fig5}
\end{figure}

Classification tasks are useful to test the nonlinear capabilities of the system as they require a higher dimensional mapping space of the input as well as some degree of memory. In this task, the TDRC must be able to correctly classify square and sine waveforms. The input sequence $u(n)$ consists of randomly ordered sequences of sine and square waves that are discretized over 12 points per period. The masking sequence $m(n)$ is uniformly distributed within the range [0, +1]. An accurate classification of the input signal must result in a predicted value of 1.0 if the signal is a square waveform, and a predicted value of 0.0 if the signal is a sine waveform. We measure the performance with the classification accuracy of the TDRC, which is calculated by dividing the number of accurate predictions by the total number. We use a training set of 2000 samples and a testing set of 1000 samples. The results in terms of the accuracy of the prediction vs the variations of $\tau_{\textrm {FC}}$ are shown in \cref{fig5}.

\subsubsection{Wireless channel equalization}

Another task of interest in RC is the capability of channel equalization in telecommunications. In the following task, a signal has been propagated through a wireless channel disturbed by noise. The channel is modeled as a linear system that is also affected by a combination of second-order and third-order nonlinear distortions due to multi-path propagation. For this task, the original signal $d(n)$ is an independent, identically distributed sequence of values $\{3, -1, +1, +3\}$. The output of the linear wireless channel model $q(n)$ is expressed as follows:

\begin{equation}\label{eq8}
\begin{split}
q(n) = 0.08d(n+2) - 0.12d(n+1) + d(n)  + 0.18d(n-1)  - 0.1d(n-2) \\ + 0.091d(n-3) - 0.05d(n-4) + 0.04d(n-5) + 0.03d(n-6)  + 0.01d(n-7).
\end{split}
\end{equation}

To generate  the input sequence of the system, $u(n)$ we take the output of the channel $q(n)$ after it has been impacted by pseudo-random additive Gaussian noise with zero mean, which we note as $v(n)$. $u(n)$ also includes the higher-order nonlinear distortions and it is defined as:

\begin{equation}
    u(n) = q(n) + 0.036q(n)^2 - 0.011q(n)^3 + v(n).
\end{equation}\label{eq9}

\begin{figure}[h!]
\centering
    \includegraphics[scale=0.615, trim={0.0cm 12.0cm 0cm 4.2cm}]{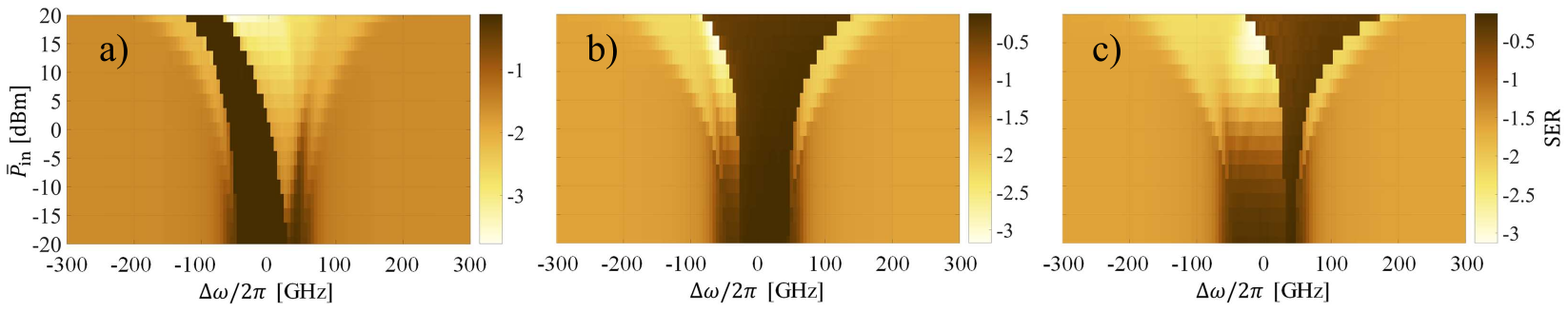}
\captionof{figure}{SER of the channel equalization task as a function of $\overline{P}_{\textrm {in}}$ and $\Delta\omega/2\pi$ for a $\tau_{\textrm {FC}}$ value of: a) 10 ps, ${\textrm b)\ 10\ \textrm{ns}}$, c) 25 ns. Minimum SER = 1.7$\times10^{-4}$ at $\tau_{\textrm {FC}}$ = 10 ps, $\overline{P}_{\textrm {in}}$ = 20 dBm, $\Delta\omega/2\pi$ = -55 GHz.}
\label{fig6}
\end{figure}

In the preprocessing of the input, we introduce a bias: $u(n) + 5$, before masking the input sequence. The masking sequence $m(n)$ is generated from a uniform distribution over the interval $[-1, +1]$. The training stage of the TDRC minimizes the square error between the reconstructed and original signals. At the post-processing stage, the output of the RC is approximated to the closest symbol in $\{3, -1, +1, +3\}$, which generates the reconstructed signal $\hat{y}(n)$. As performance metric, we calculate the symbol error ratio (SER) between the original and reconstructed signals at a signal-to-noise ratio (SNR) of 32 dB. The obtained results as a function of $\overline{P}_{\textrm {in}}$ and $\Delta\omega$ are shown in \cref{fig6} for increasing values of $\tau_{\textrm{FC}}$ in base-10 logarithm scale. 10000 data samples are used for training and 10 different subsets of 10000 samples for testing. 
\newpage
\vspace{-0.2cm}
\subsection{Memory capacity vs free-carrier lifetime}\label{subsec:4.2}

Now, we focus on obtaining the variation of the memory capacity over the parameter space defined in terms of $\overline{P}_{\textrm {in}}$ and $\Delta\omega/2\pi$ as we did with the performance metric of the task while varying the value of $\tau_{\textrm {FC}}$. It is necessary to set an upper limit to the order of memory capacity that is going to be calculated. Therefore, based on the mathematical expressions that define each of the tasks, we have considered the nonlinear memory capacity up to the $3^{\textrm{rd}}$ order to have the most significance in this study. We use the same input sequence as for the NARMA-10 task to calculate $MC$.

\begin{figure}[h!]
\centering
    \includegraphics[scale=0.625, trim={0.2cm 12.0cm 0cm 4.2cm}]{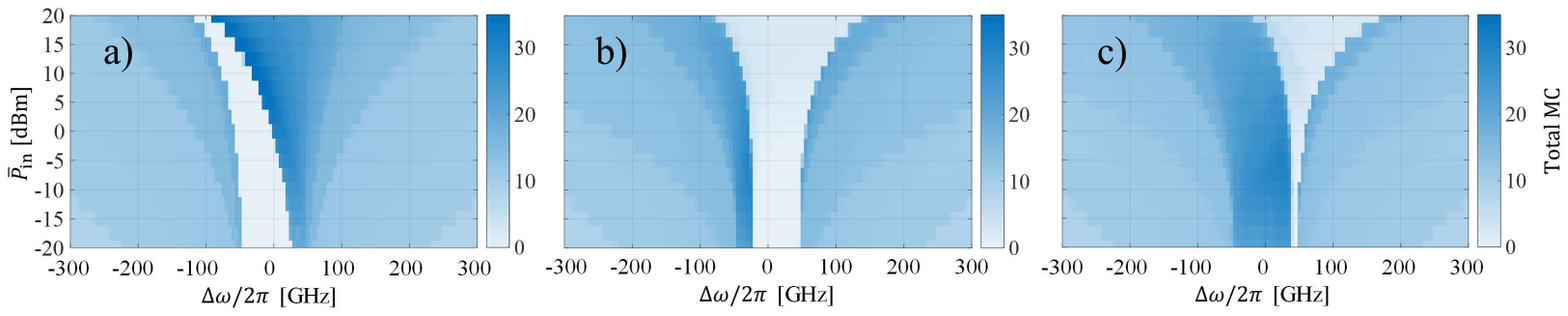}
\captionof{figure}{Total memory capacity (up to $3^{\textrm{rd}}$ order) of the simulated MRR-based TDRC as a function of $\overline{P}_{\textrm {in}}$ and $\Delta\omega/2\pi$ for a $\tau_{\textrm {FC}}$ value of: a) 10 ps, b) 10 ns, c) 25 ns.}
\label{fig7}
\end{figure}
\vspace{-0.2cm}

\subsection{Nonlinear detuning vs free-carrier lifetime}\label{subsec:4.3}

\begin{figure}[h!]
\centering
    \includegraphics[scale=0.62, trim={0cm 12.2cm 0cm 3.8cm}]{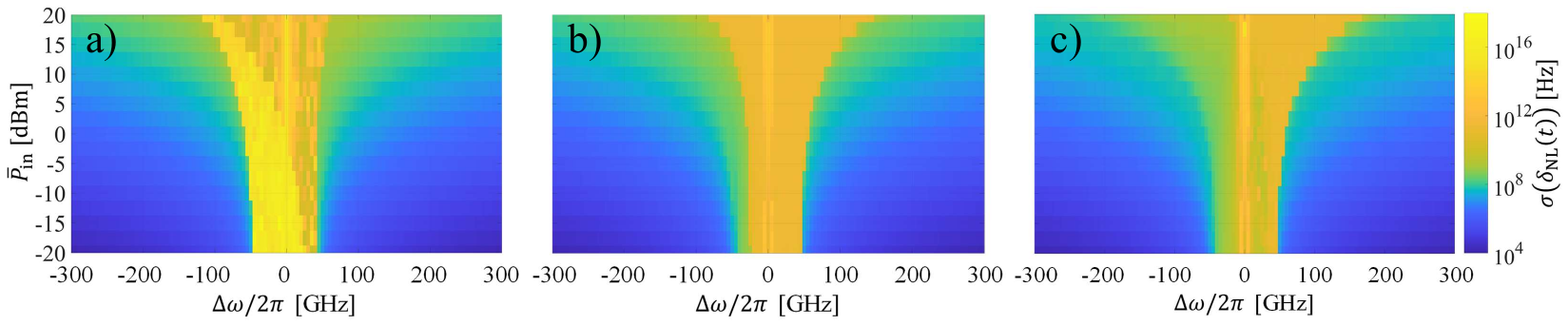}
\captionof{figure}{Nonlinear detuning in the silicon MRR cavity of the simulated MRR-based TDRC as a function of $\overline{P}_{\textrm {in}}$ and $\Delta\omega/2\pi$ for a $\tau_{\textrm {FC}}$ value of: a) 10 ps, b) 10 ns, c) 25 ns.}
\label{fig8}
\end{figure}

Similarly, we simulate the nonlinear detuning for the same range of power, frequency detuning, and set of values of $\tau_{\textrm {FC}}$ (\cref{fig8}). The results correspond to simulations addressing the NARMA-10 task. However, due to the bias added to the input sequences, it can be assumed that the differences in the value of $\sigma(\delta_{\textrm{NL}}(t))$ between the different tasks are negligible. Hence, we assume $\sigma(\delta_{\textrm{NL}}(t))$ to have the same value in every task at a given $\overline{P}_{\textrm {in}}$ and $\Delta\omega/2\pi$. Such bias is a requirement of the mathematical model of the system to generate a quasimonochromatic optical input signal (small amplitude variations) \cite{GironCastro:24}. Experimentally, this assumption for $\sigma(\delta_{\textrm{NL}}(t))$ would not necessarily hold depending on the optical modulation procedure of each task.

\subsection{Memory capacity vs performance and nonlinear detuning}\label{subsec:4.4}

In this subsection, we compare the task-independent metric of the memory capacity with the performance of the NARMA-10 (0 dBm), signal classification (0 dBm), radar (0 dBm, $k = 2$), and channel equalization ($20\ \textrm{dBm}$) tasks. The analysis is quantified in terms of $\Delta\omega$ for a single value of $\overline{P}_{\textrm {in}} $ = 0 dBm and different values of $\tau_{\textrm {FC}}$. This comparison in a 2-D frequency sweep offers a clearer picture of the relation between each performance metric and the memory capacity.

\begin{figure}[h!]
\centering
    \includegraphics[scale=0.4610, trim={0cm 0cm 0cm 0cm}]{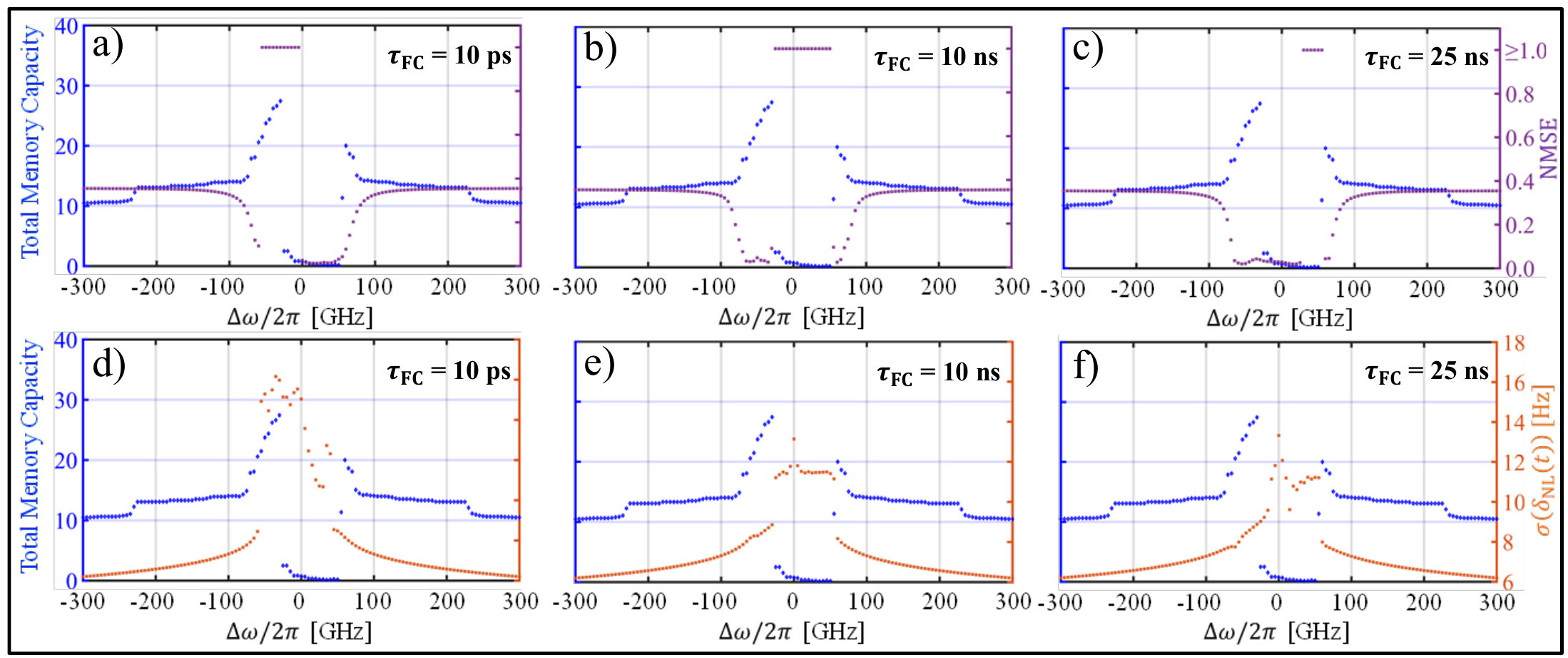}
\captionof{figure}{NARMA-10 task: Prediction NMSE (a - c), nonlinear detuning (d - f) and total memory capacity as a function of $\Delta\omega/2\pi$ for a $\overline{P}_{\textrm {in}}$ = 0 dBm and $\tau_{\textrm {FC}}$: 10 ps (a, d), 10 ns (b, e), and 25 ns (c, f).}
\label{fig9}
\end{figure}

\begin{figure}[h!]
\centering
    \includegraphics[scale=0.4610, trim={0cm 0cm 0cm 0cm}]{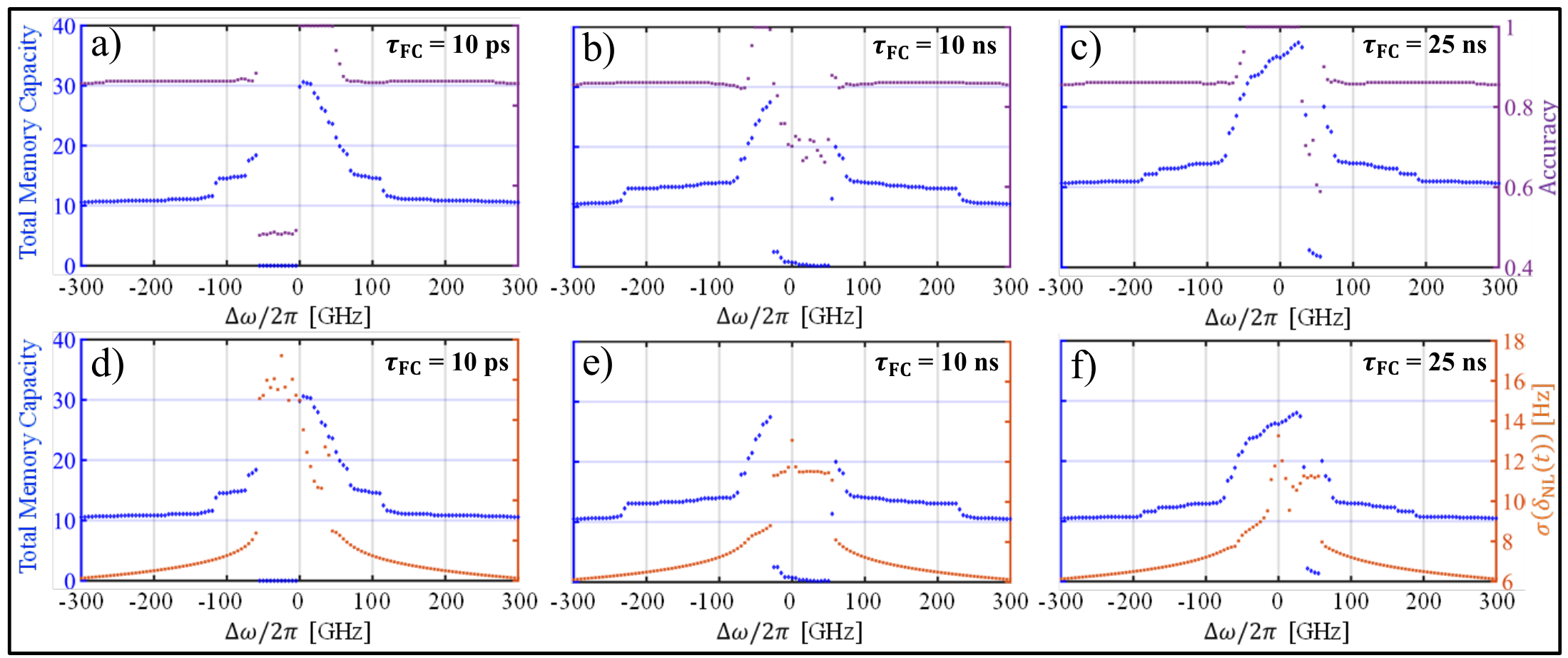}
\captionof{figure}{Signal classification task: Prediction accuracy (a - c), nonlinear detuning (d - f) and total memory capacity as a function of $\Delta\omega/2\pi$ for a $\overline{P}_{\textrm {in}}$ = 0 dBm and $\tau_{\textrm {FC}}$: 10 ps (a, d), 10 ns (b, e), and 25 ns (c, f).}
\label{fig10}
\end{figure}

\begin{figure}[h!]
\centering
    \includegraphics[scale=0.4612, trim={0cm 0cm 0cm 0cm}]{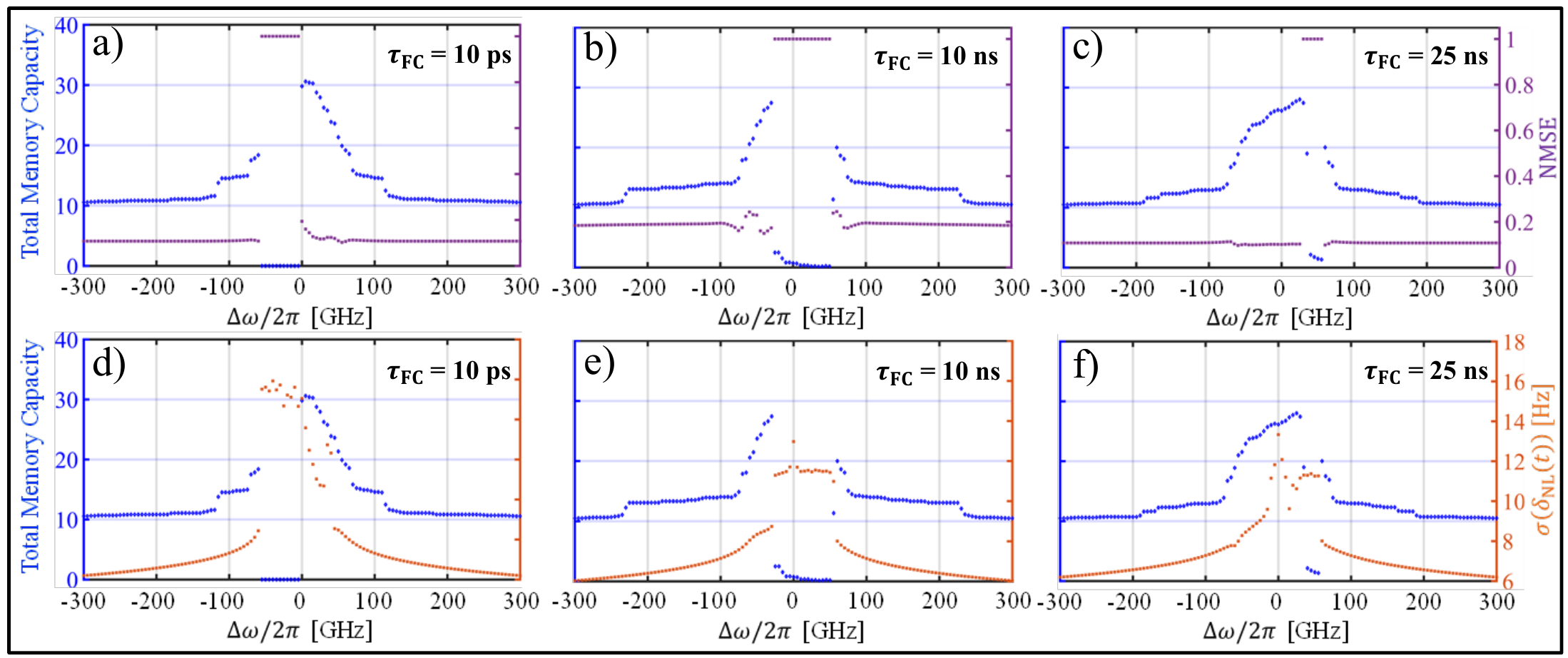}
\captionof{figure}{Radar task: Prediction NMSE (a - c), nonlinear detuning (d - f) and total memory capacity as a function of $\Delta\omega/2\pi$ for$\overline{P}_{\textrm {in}}$ = 0 dBm, $k = 2$ and  $\tau_{\textrm {FC}}$: 10 ps (a, d), 10 ns (b, e), and 25 ns (c, f).}
\label{fig11}
\end{figure}

\begin{figure}[h!]
\centering
    \includegraphics[scale=0.461, trim={0cm 0cm 0cm 0cm}]{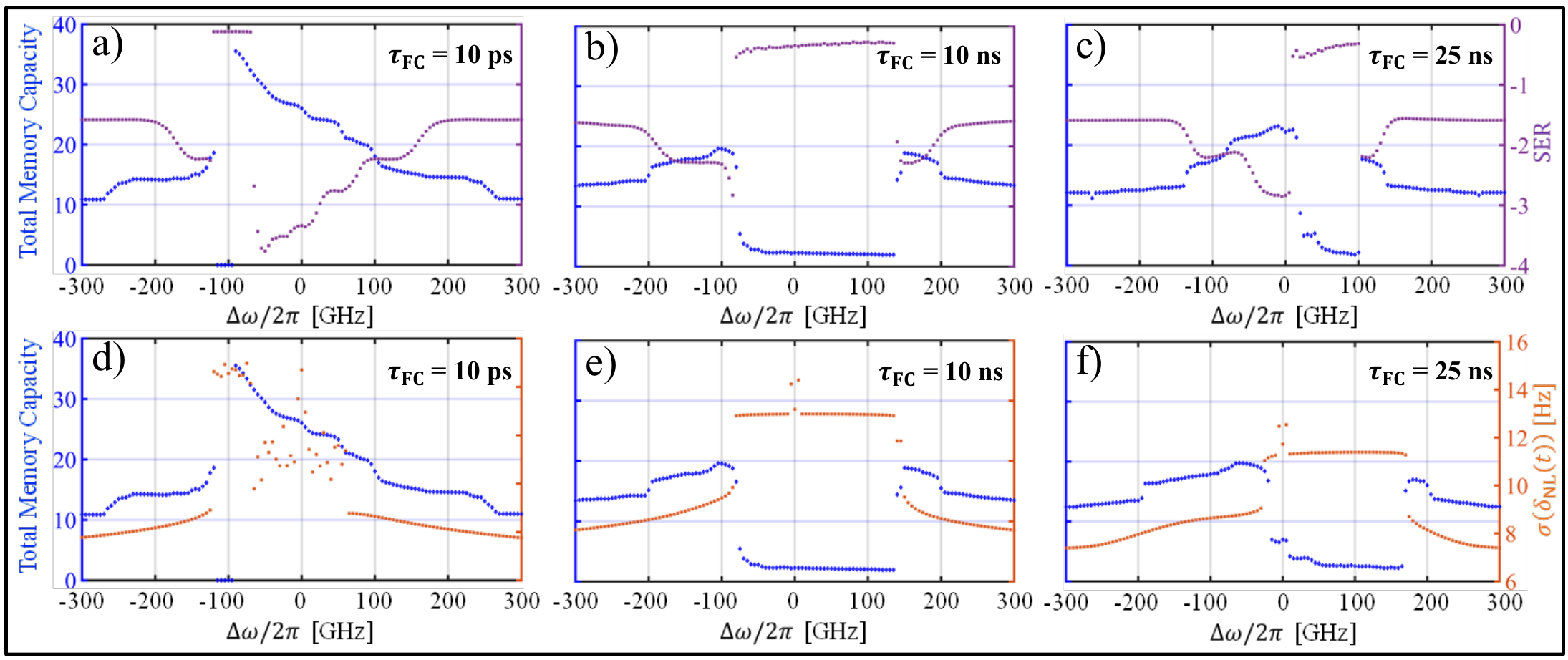}
\captionof{figure}{Channel equalization task: Prediction SER (a - c), nonlinear detuning (d - f) and total memory capacity as a function of $\Delta\omega/2\pi$ for $\overline{P}_{\textrm {in}}$ = 20 dBm and $\tau_{\textrm {FC}}$: 10 ps (a, d), 10 ns (b, e), and 25 ns (c, f).}
\label{fig12}

\end{figure}

\newpage
\section{Discussion}\label{sec:5}

The results corresponding to the prediction performance of each task (\cref{fig3,fig4,fig5,fig6}) demonstrate that the regions (A, B, C) of the parameter space defined by $\Delta\omega/2\pi$ for $\overline{P}_{\textrm {in}}$, generally persist, even among tasks of different computing nature and applications. Region C achieves the highest level of nonlinear detuning as shown in \cref{fig8}. This detuning is sufficient to set the resonance of the MRR too far from the optical signal frequency, pushing the cavity into a self-pulsing behaviour, out of resonance. As the nonlinear detuning for a specific value of $\Delta\omega/2\pi$ and $\overline{P}_{\textrm{in}}$ remains constant from task to task, it is not surprising that the self-pulsing behaviour is located in the same parameter space for every task. Hence, the large performance penalty of region C is distributed over the same range of $\Delta\omega/2\pi$ and $\overline{P}_{\textrm{in}}$ for every task and represents the lowest performance (\cref{fig3,fig4,fig5,fig6}) and the lowest memory capacity (\cref{fig7}) achieved. Self-pulsing harms heavily the memory due to its characteristic disrupting behaviour which prevents the TDRC from maintaining consistency at the training\cite{GironCastro:24}. Likewise, all the tasks converge to the best performance value in a fully linear regime as $|\Delta\omega/2\pi|$ increases while $\sigma(\delta_{\textrm{NL}}(t))$ decreases towards the order of tens of kHz (\cref{fig8}). In other words, the MRR operates in region A. As aforementioned, this performance of the TDRC in region A is achievable thanks to the intrinsic nonlinearity of the system at photodetection. Region A also benefits from a moderate memory capacity as shown in \cref{fig7}. 

Nevertheless, some important features of this system are highlighted by the difference between the tasks in the extension of region B (best performance of the system) over the parameter space. We take the results of the NARMA-10 task (\cref{fig3}) as a reference for comparison. For the NARMA-10 task, region B is characterized by a moderate level of nonlinear detuning, in the order of hundreds of MHz  (\cref{fig3,fig8,fig10}). Region B spans over both negative detuning (dominance of the TO effect) and positive detuning (dominance of FCD). \cref{fig7} shows that region B has the highest amount of memory capacity as long as self-pulsing is not triggered \cite{GironCastro:24}. However, when less memory is required, like the radar task, the distinction between the performances of region A and B is not as significant \cref{fig4}(a - c) as the memory capacity in the linear regime appears to be enough to achieve good performance. The performance difference between different levels of nonlinearity becomes more clear when the prediction time horizon is increased $(k = 2)$ \cref{fig4,fig11}(d - f), although it is not comparable to the clear difference of performance in the NARMA-10 task.  On the other hand, the signal classification task appears to require a higher nonlinear detuning (in the order of GHz as shown in \cref{fig10}) with respect to the NARMA-10 task. This difference of one order of magnitude narrows the size of region B over the parameter space. This requirement of a higher level of nonlinearity is to be expected due to the classification nature of the task. 

The wireless channel equalization task involves the most different requirements of nonlinearity and memory in comparison to the rest of the tasks. In fact, it is hard to establish a clear picture of the right properties of the system to achieve good performance in this task. As shown in \cref{fig6}, it definitely benefits from higher levels of power, unlike the other tasks, which under certain conditions, can achieve good performance at lower power as well. Nonetheless, the span of the parameter space of low SER is very narrow for this system as it is limited to a small set of values of $\Delta\omega/2\pi$ and $\overline{P}_{\textrm{in}}$. As shown in \cref{fig12} the performance of this task seems to be the best in the limited range of values where very high nonlinear detuning (1$\times10^{10}$ - 1$\times10^{14}$ Hz) and high memory capacity are able to coexist. However, this often happens when the MRR is on the verge of self-pulsing, which penalizes the memory capacity if triggered, making region B considerably narrower than in the rest of the tasks. The channel equalization and radar tasks, which involve signals affected by noise and nonlinear distortions, appear to narrow the extension of region B if compared to that of other tasks. Particularly, within the range $\Delta\omega/2\pi \approx [-60, 60]$ GHz with $\overline{P}_{\textrm{in}} \approx [-20, 0]$ dBm (not taking into account the parameter space where region C is present). The nonlinear detuning appears to be insufficient to achieve proper training of the system, even if the memory capacity is high in that range. Consequently, the performance becomes even worse than in the linear regime, and comparable to the lowest performance due to self-pulsing in region C. As previously studied\cite{HülserKöster2023}, this poor correlation between memory capacity and performance may occur in TDRC under certain conditions of the system and tasks.

As in our previous study\cite{GironCastro:24}, the variation of $\tau_{\textrm {FC}}$ provides the potential of improving the performance of each of the tasks, just like it was demonstrated initially with the NARMA-10 task. As the results also demonstrate in this work (\cref{fig7}), the maximum memory capacity can also be increased by decreasing $\tau_{\textrm {FC}}$, and the size of region B can be extended by mitigating the self-pulsing, which also extends the parameter space where high memory capacity is present. Therefore, a reconfigurable implementation of this system in which control of the cavity free carrier lifetime is possible e.g., with a PIN junction, in addition to the control of input power and frequency detuning, would potentially make the photonic TDRC capable of handling different applications.

\section{Conclusion}\label{conclusion}

Throughout this work, we have analyzed the performance, memory capacity, and nonlinearity properties of a silicon MRR-based TDRC. The results provided insight into the specific differences in the computing nature of each task. These differences, determine the requirement of input power and frequency detuning of each task within the region of adequate nonlinearity provided by the MRR. We also have validated that the impact of the free carrier lifetime on the defined regions is extrapolated to other tasks different from NARMA-10. This extrapolation is particularly evident in the results regarding memory capacity and nonlinear detuning, which are task-independent under the established simulation conditions.  Our work allows us to further generalize the beneficial or detrimental impact of self-pulsing, nonlinear, and linear regimes of an MRR on tasks covering a diversity of applications.

\appendix    %>>>> this command starts appendixes

\acknowledgments % equivalent to \section*{ACKNOWLEDGMENTS}       
This work has received funding by Vetenskapsrådet (BRAIN, grant n. 2022-04798) and Villum Founden (OPTIC-AI VIL29334, and VI-POPCOM 54486).

\bibliography{report} % bibliography data in report.bib

\begin{thebibliography}{10}

\bibitem{Schuman2022}
Schuman, C.~D., Kulkarni, S.~R., Parsa, M., Mitchell, J.~P., Date, P., and Kay, B., ``Opportunities for neuromorphic computing algorithms and applications,'' {\em Nat. Comput. Sci.}~{\bf 2},  10--19 (Jan 2022).

\bibitem{Huang2022}
Huang, C., Sorger, V.~J., Miscuglio, M., Al-Qadasi, M., Mukherjee, A., Lampe, L., Nichols, M., Tait, A.~N., de~Lima, T.~F., Marquez, B.~A., Wang, J., Chrostowski, L., Fok, M.~P., Brunner, D., Fan, S., Shekhar, S., Prucnal, P.~R., and Shastri, B.~J., ``Prospects and applications of photonic neural networks,'' {\em Adv. Phys. X}~{\bf 7} (Oct 2022).

\bibitem{Cucchi_2022}
Cucchi, M., Abreu, S., Ciccone, G., Brunner, D., and Kleemann, H., ``Hands-on reservoir computing: a tutorial for practical implementation,'' {\em Neuromorphic Comput. Eng.}~{\bf 2},  032002 (Aug 2022).

\bibitem{jaeger2001short}
Jaeger, H., ``Short term memory in echo state networks,'' {\em GMD Rep.}~{\bf 152},  60 (May 2002).

\bibitem{Paquot2012}
Paquot, Y., Duport, F., Smerieri, A., Dambre, J., Schrauwen, B., Haelterman, M., and Massar, S., ``Optoelectronic reservoir computing,'' {\em Sci. Rep.}~{\bf 2},  287 (Feb 2012).

\bibitem{Bueno:17}
Bueno, J., Brunner, D., Soriano, M.~C., and Fischer, I., ``Conditions for reservoir computing performance using semiconductor lasers with delayed optical feedback,'' {\em Opt. Express}~{\bf 25},  2401--2412 (Feb 2017).

\bibitem{8758193}
Röhm, A., Jaurigue, L., and Lüdge, K., ``Reservoir computing using laser networks,'' {\em IEEE J. Sel. Top. Quantum Electron.}~{\bf 26},  1--8 (Jan 2020).

\bibitem{Skontranis_2023}
Skontranis, M., Sarantoglou, G., Sozos, K., Kamalakis, T., Mesaritakis, C., and Bogris, A., ``Multimode fabry-perot laser as a reservoir computing and extreme learning machine photonic accelerator,'' {\em Neuromorphic Comput. Eng.}~{\bf 3},  044003 (Oct 2023).

\bibitem{Duport:12}
Duport, F., Schneider, B., Smerieri, A., Haelterman, M., and Massar, S., ``All-optical reservoir computing,'' {\em Opt. Express}~{\bf 20},  22783--22795 (Sep 2012).

\bibitem{doi:10.34133/icomputing.0067}
Biasi, S., Donati, G., Lugnan, A., Mancinelli, M., Staffoli, E., and Pavesi, L., ``Photonic neural networks based on integrated silicon microresonators,'' {\em Intelligent Computing}~{\bf 3},  0067 (Jan 2024).

\bibitem{Chen:19}
Chen, Y., Yi, L., Ke, J., Yang, Z., Yang, Y., Huang, L., Zhuge, Q., and Hu, W., ``Reservoir computing system with double optoelectronic feedback loops,'' {\em Opt. Express}~{\bf 27},  27431--27440 (Sep 2019).

\bibitem{GironCastro:24}
{Giron Castro}, B.~J., Peucheret, C., Zibar, D., and {Da Ros}, F., ``Effects of cavity nonlinearities and linear losses on silicon microring-based reservoir computing,'' {\em Opt. Express}~{\bf 32},  2039--2057 (Jan 2024).

\bibitem{PhysRevA.87.053805}
Zhang, L., Fei, Y., Cao, T., Cao, Y., Xu, Q., and Chen, S., ``Multibistability and self-pulsation in nonlinear high-{$Q$} silicon microring resonators considering thermo-optical effect,'' {\em Phys. Rev. A}~{\bf 87},  053805 (May 2013).

\bibitem{Inubushi2017}
Inubushi, M. and Yoshimura, K., ``Reservoir computing beyond memory-nonlinearity trade-off,'' {\em Sci. Rep.}~{\bf 7},  10199 (Aug 2017).

\bibitem{Dambre2012}
Dambre, J., Verstraeten, D., Schrauwen, B., and Massar, S., ``Information processing capacity of dynamical systems,'' {\em Sci. Rep.}~{\bf 2},  514 (Jul 2012).

\bibitem{HülserKöster2023}
Hülser, T., Köster, F., Lüdge, K., and Jaurigue, L., ``Deriving task specific performance from the information processing capacity of a reservoir computer,'' {\em Nanophotonics}~{\bf 12},  937--947 (Oct 2023).

\bibitem{Haykin}
Haykin, S.
\newblock The Dartmouth database of IPIX radar (2001) http://soma.ece.mcmaster.ca/ipix/dartmouth/ind ex.html Last time accessed on 20 March 2024.

\end{thebibliography}
\bibliographystyle{spiebib} % makes bibtex use spiebib.bst

\end{document}